# From product to system network challenges in system of systems lifecycle management

Vahid Salehi
salehi-d@hm.edu
Munich University of Applied Sciences
Munich, German

Shirui Wang
Engineering Design of Mechatronic Systems
Munich, Germany
s.wang@edmrc.de

Josef Vilsmeier
Engineering Design of Mechatronic Systems
Munich, Germany
j.vilsmeier@edmrc.de

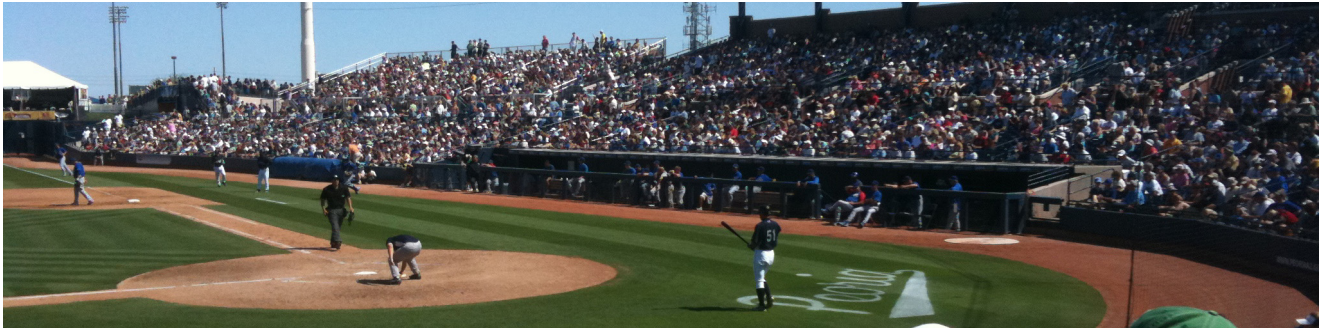

Figure 1: Article in the anniversary issue of the journal EDMS e.V.2025

## Abstract

Today, products are no longer isolated artifacts, but nodes in networked systems. This means that traditional, linearly conceived life cycle models are reaching their limits: Interoperability across disciplines, variant and configuration management, traceability, and governance across organizational boundaries are becoming key factors. This collective contribution classifies the state of the art and proposes a practical frame of reference for SoS lifecycle management, model-based systems engineering (MBSE) as the semantic backbone, product lifecycle management (PLM) as the governance and configuration level, CAD-CAE as model-derived domains, and digital thread and digital twin as continuous feedback. Based on current literature and industry experience, mobility, healthcare, and the public sector, we identify four principles: (1) referenced architecture and data models, (2) end-to-end configuration sovereignty instead of tool silos, (3) curated models with clear review gates, and (4) measurable value contributions along time, quality, cost, and sustainability. A three-step roadmap shows the transition from product- to network-centric development: piloting with reference architecture, scaling across variant and supply chain spaces, organizational anchoring (roles, training, compliance). The results are increased change robustness, shorter throughput times, improved reuse, and informed sustainability decisions. This article is aimed at decision-makers and practitioners who want to make complexity manageable and design SoS value streams to be scalable.

## 1 Introduction

More and more products today are no longer isolated units, but part of a larger network of technical, organisational and digital systems. This development calls for a rethink: classic lifecycle approaches are reaching their limits because dependencies and interactions are increasing. Managing such system networks requires new methods to ensure transparency, traceability and adaptability. The ability to collaborate across disciplines and domains from mechanics to electronics to software is central to this. This article shows how companies are facing this change, what obstacles they encounter and what principles help to make complexity manageable. The integration of CAD, PLM and systems of systems is becoming increasingly important. Introduction: Nowadays, what we used to call "a product" is no longer a self-contained object. Rather, many offerings are understood as part of larger systems, which in turn can be combined into even larger networks so-called systems of systems (SoS). This development places the lifecycle management of such networks at the centre of a new challenge. Whereas in the past the focus was on a machine, a component or a piece of software, today the focus is on how mechanical components, electronics, software, data flows and services can be integrated across disciplines and domains. The classic product lifecycle design, development, production, operation, disposal is no longer sufficient. Normative guidelines such as the ISO IEC IEEE 15288 standard provide a solid basis, describing all phases from design to use to decommissioning [16]. Nevertheless, it is clear that these lifecycle processes alone are no longer sufficient for a system of systems scenario. Environments have emerged in which operational independence of subsystems, evolutionary development, heterogeneous components and interfaces across domains are part of everyday life. As a result, our understanding of lifecycle management must evolve. When an industrial company is confronted with the question of how to manage the design of a product that is embedded in a larger system, many issues arise: How can CAD models, PDM PLM data, requirements, system architectures and service data be linked consistently? How can requirements be tracked right through to operation? What governance applies when several partners from different domains are involved? And last but not least: How do you keep track of variants, feedback and dynamics in operation? All these questions show that lifecycle management in the SoS context is not an organisational add-on it is the core discipline of modern product and system development. A central aspect is the demand for model-based systems engineering (MBSE). Models are no longer understood as a secondary framework, but as a binding artefact that describes requirements, system behaviour, interfaces and structures. Studies have shown that models can contribute significantly to improved communication and highlight interactions between subsystems [3]. In an SoS environment, this is where the added value lies: when a subsystem operates independently but interacts with others, emergent properties can arise properties that cannot be identified by analysing the individual parts alone. This is where MBSE helps to describe the bigger picture, identify the right interfaces and reveal interdependencies at an early stage. But MBSE alone is not enough. A governance layer is needed that works across different data types, tools and organisational units. This is exactly where product lifecycle management (PLM) comes into play. Previously perhaps seen only as a document or parts list container, PLM has evolved into a platform that controls variant management, change processes, configuration sovereignty and the integration of specialist disciplines. In the SoS context, PLM is no longer a nice-to-have it is essential to ensure that CAD geometries, system architectures and operating data, for example, do not end up in isolated solutions. In ETO (engineering-to-order) programmes, for example, it is clear that the coordination of PLM processes with SoS structures can make the difference between chaos and controllability [7]. But how can CAD 3D design be successfully linked to system architecture and modelling? If a company continues to work in CAD in parameters, geometries, tolerances then the question arises: how does this level get into the system model? And how can it be ensured that changes are not lost there, but are fed back into the system model and the PLM chain? Research shows that workspaces, parametric design and functional geometries can now serve as a bridge to integrate CAD artefacts into a system architecture [10]. Design is no longer thought of in isolation, but as part of a larger whole as a module that integrates with other parts of the system. Another crucial idea: digital twins and digital threads. When we link operating and service data with design and architecture, a feedback loop is created a continuous data thread throughout the lifecycle. This not only maps the status of a system network, but also allows it to be actively controlled, variants to be analysed and feedback loops to be automated. A study on System of Systems Lifecycle Engineering highlights how smart products and services are now becoming part of a larger SoS and what an SoS lifecycle concept must look like in order to enable such feedback loops [1]. But we are far from having overcome all the challenges. One of the biggest can be summed up with the word "adoption." As many organisations are now realising, methods and tools alone are not enough. Roles, culture, data sovereignty, partner networks, interface standards all of these must be taken into account. Studies on the introduction of MBSE reveal constant hurdles: complexity, lack of compatibility with established processes and few tangible use cases mean that many companies are only half-heartedly introducing MBSE SoS-LCM [23], [29]. An effective remedy here is the action research approach: industry and research work together on pilot projects, analyse, control and learn together and this results in methods that work in practice and not just in science. The dimension of sustainability must not be forgotten either. In the context of SoS lifecycle management, it is no longer just about getting to market as quickly as possible or keeping costs low. Increasingly, it is about ecological and social aspects: material use, reparability, recyclability, service life. The literature shows that, especially in the case of systemic products and networks, sustainability cannot be considered downstream. Rather, it should be anchored in the model and in the data governance system from the outset [9], [13]. This in turn requires that design, architecture and operation are not considered in isolation from one another. In the end, the question remains: How can we design a contribution for manufacturers, service providers and operators that does not end in abstract sets of methods, but rather leads to clear recommendations for action? My suggestion is to start with a reference use case that exemplifies how a product is not created in isolation but is embedded with CAD PLM MBSE tools and a digital thread from deployment to disposal. From the outset, build an architecture and model governance that is not understood as a superstructure but as an integral part of the development process. Ensure that the design is not "maintained" after the fact, but is conceived as part of the model in conjunction with the system architecture and in configuration via PLM. At the same time, establish a digital twin that is not only created during operation, but is also considered during development. And don't forget the organisational side: clearly define roles, regulate interfaces and data sovereignty, carry out pilot projects with clear objectives, and plan learning cycles. This is how you can successfully manage change: from isolated products to networked system families, from linear life cycles to networked lifecycle chains, from separate domains to interdisciplinary cooperation. This allows you to embrace complexity rather than suppress it and, if necessary, to benefit from it: greater flexibility, shorter response times to changes, better feedback from operations and more sustainable product system offerings. The path is not easy; it requires courage, investment in processes and data, and cultural change but it is unavoidable if you want to not only survive but thrive in the age of system networks.

## 2 Literature research: System of Systems Lifecycle Management (SoS-LCM) in interaction with CAD, PDM PLM, MBSE, 3D design and PEP

Companies are increasingly moving away from isolated products to system networks consisting of mechanical, electrical, software, and service-based subsystems. This "system of systems" (SoS) has characteristics such as operational and organizational independence of the subsystems, emergent behavior, and evolutionary development pushing classic linearly conceived life cycle models to their limits [1][4], [15], [17]. The ISO IEC IEEE family of standards provides the process basis for this: 15288 defines the system life cycle processes, 24748-1 provides practical guidelines for adaptation in domains, and 1471 (now 42010) sets the framework for architecture descriptions and views [15][17], [21]. These standards form the connecting foundation for aligning MBSE, PLM PDM, CAD 3D design, and PEP particularly important when SoS come from multiple organizations and data sovereignty, interfaces, variants, and changes need to be synchronized [3], [7], [11], [12], [15].

### 2.1 Why SoS lifecycle management is different

In SoS, dependencies and interactions arise across system boundaries. It is not enough to simply version your "own" product; end-to-end transparency is required across requirements, architecture, behavior, interfaces, and configurations ideally model-based right through to operation and service. Research from Cambridge (Design Science, Proceedings of the Design Society) and university repositories shows that MBSE acts as a common language and "bracket" between disciplines; PLM provides governance, data continuity, and variant logic; CAD CAE feeds geometry parameters; PDM organizes design results; DT DI (digital twin threads) extend the chain to operation feedback [1][4], [6][14], [19], [22]. This is both a technical and organizational task: roles, processes, toolchains, and data sovereignty must fit together [3], [7], [8], [12]. MBSE as the backbone and its maturity The literature paints a clear picture: MBSE improves traceability, consistency, and reusability, but often faces adoption hurdles (complexity, tool compatibility, lack of reference models) [18], [23], [24], [27], [29]. Current work differentiates between types of use (userisation of system models) and calls for targeted reuse transformation of system models into domain models (e.g. CAD simulation) instead of duplicate maintenance [3]. At the same time, work is being done on language and ontology issues: further development of SysML, LML as a complementary ontology, as well as guidelines for model curation and human centered, interactive model environments (IMCSE) [5], [19], [28], [30], [31]. These strands are central to reliably anchoring traceability across disciplines (requirements to architecture to design to verification to operation) a core objective in SoS-LCM [3], [11], [15]. PLM PDM as a governance layer beyond a "data container".PLM is evolving from a pure document bill of materials container to a knowledge and governance platform across the lifecycle: process views for engineering change, configuration variable management, rights sovereignty, lifecycle status and approvals including links to CAD CAE, requirements tools, ALM software repositories, and operating systems (IoT service) [7], [10], [12], [25]. Process guidelines for PLM implementation are particularly crucial in engineering-to-order (ETO) programs (Cambridge Design Science; McKendry et al.), as any deviation becomes exponentially more expensive in later phases [7], [12]. In addition, Cambridge's work on sustainability in PLM emphasizes that ecological social criteria should be incorporated into decisions at an early stage in a model- and data-driven manner, an increasingly mandatory field in the SoS context [9]. 5. CAD 3D design and system architecture bridge instead of island A recurring finding: CAD artifacts must not only be linked to the system model, but also derived from it and fed back into it (parameters, working surfaces, functional surfaces, tolerances). Cambridge contributions show how parametric working surfaces and contact channel methods can be integrated into MBSE-supported PLM approaches to consistently implement changes [10]. In modular product families (e.g. vehicle construction), MBSE in modular development makes cross-relationships manageable because architecture variance decisions are explicitly modeled and synchronized with CAD CAE [6].

### 2.2 Digital twins and the "common thread"

Digital twins only function reliably in SoS when they are defined on a model basis and supplied with lifecycle data: MBSE provides structure behavior interfaces , PLM provides valid configurations, and telemetry operating data close the loop. The literature (including NPS, MIT author versions) sees MBSE as a way to start DT development earlier in the life cycle and harmonize data flows [20], [22]. In Cambridge contexts and more recent work, the coupling of PLM and digital twins is also evident in "product generation development" and along the digital thread [4], [11]. Integration of PEP process, from requirements to testing. For PEP (product development process), this means that requirements → architecture → behavior → design (CAD electrical software), → simulation verification → integration → operation must be connected in a model-driven and data-driven manner. Normative processes (15288 24782-1) can be adapted for use in this context; they are not an end in themselves, but rather a framework for continuous reviews, model gateways, and configured artifacts [15], [16], [21]. The test verification documentation (IEEE guidelines) is anchored in the model tests are linked to requirements architecture and controlled via PLM ALM [26]. This creates traceability and robustness to changes; exactly what makes changes to SoS (new interface, modified firmware, new supplier sensor) absorbable over lifecycles [3], [11], [15].

### 2.3 Adoption hurdles and success factors

Implementation rarely fails because of theory, but rather because of roles, culture, tool landscape, and training. Surveys and case studies (including Call 2024; Pandolf 2023; IMCSE MIT) show that perceived complexity, lack of compatibility with existing practices and a lack of reference models examples slow progress [23], [29][31]. Successful programs take a domain-specific approach (e.g. healthcare construction, navy DoD, aviation), start with pilot use cases and reference architectures, clearly define ownership governance via PLM and establish model curation (naming, packages, reviews) as an independent discipline [7], [12], [28], [31], [33]. Action research approaches (industry to research) have proven effective in making methods practical and getting organizations on board [34], [35].

## 2.4 Sustainability, circularity and compliance in SoS

New work anchors sustainability criteria in PLM MBSE: material and energy aspects, reparability, second life, circular strategies, and regulatory requirements (e.g. documentation obligations) are to be recorded in a model-based manner and evaluated along the digital thread of the [9], [36]. Older Cambridge articles refer early on to the shift from "waste management" to genuine lifecycle management, now more relevant than ever in SoS [13]. What does this mean specifically for your anthology contribution No. 1? Based on this literature, your editorial can make four things clear:

1. Why classic lifecycle thinking is no longer sufficient: SoS characteristics, distributed ownership, emergence with standards foundation 15288 24748-1 1471 as the common thread [1], [15][17].

2. How MBSE+PLM bridge the gap: Model-based requirements architectures as a "single source of truth"; PLM as the governance configuration backbone; CAD 3D as a derived feedback view [3], [6][8], [10][12].

3. How the digital twin belongs in the lifecycle: early modeling, providing reliable configuration and operating data [4], [11], [22].

4. How to get started organizationally: clear ownership, referenced use cases, piloting, model curation, training flanked by action research with university partners [7], [12], [28], [31], [34], [35].

Brief summary of your contribution ""

- Thesis 1: SoS-LCM requires a model and data-driven bracket, MBSE for structure behavior interfaces, PLM for governance configuration; CAD 3D and simulation are synchronized instead of being maintained in isolation [3], [6][8], [10][12], [15], [16].
- Thesis 2: Standards provide stability during adaptation: structure roles, milestones, artifacts, and views independently of tool brands [15][17], [21], [26].
- Thesis 3: Digital twins are effective when based on system models and PLM configurations; this is the only way to ensure consistency in changes, variants, and field data [4], [11], [20], [22].
- Thesis 4: Adoption succeeds with pilot cases, reference architectures, model curation, and action research loops; training and governance are just as important as tools [7], [12], [28], [31], [34], [35].

endverbatim

## 3 Conclusion

The continuous development of parametric, associative CAD systems and their integration into industrial processes has become a key research focus over the last two decades. In early work, Salehi and McMahon emphasized the importance of action research in the industrial application of such systems to gain practical insight into increasing efficiency [43]. Based on this, a generic integrated approach for parametric associative CAD systems was developed that improves the adaptability of complex product structures throughout the entire life cycle [44]. The methodical integration of such CAD systems into a PLM (product lifecycle management) environment was identified as a key component to ensure the consistency, reusability, and end-to-end traceability of technical information [45]. In the automotive sector in particular, this led to an integrated approach to powertrain design, in which parametric dependencies were systematically embedded in the development process [46]. The development of suitable evaluation frameworks for the implementation of these methods in an industrial context represented another milestone [47]. Factors such as data integration, employee acceptance, and training requirements were taken into account to enable sustainable implementation. This resulted in practice-oriented approaches that were applied in real industrial environments and demonstrated the advantages of integrated CAD linking [48]. This was also explored in depth in university research work, such as that of Salehi-Douzloo, which bridged the gap between theoretical methodology and industrial implementation [49]. In parallel, Salehi developed a System-Driven Product Development (SDPD) model that positions the integration of mechatronic systems as the core principle of product development in industry [50][51]. This system-driven methodology was presented at international PLM conferences and emphasized the connection between model-based systems engineering (MBSE) and industrial practice [52]. In later work, Salehi showed that identifying success factors in the introduction of PLM systems, in particular, organizational embedding and process integration, is crucial for the effectiveness of these methods [53]. Based on this, systems engineering approaches were embedded more deeply in PLM to design mechatronic product architectures in an integrative manner [54]. The combination of point cloud technologies with systemic development approaches opened up new perspectives for digital factory planning [55]. Together with Taha and other researchers, Salehi developed an MBSE-based IoT sensor network that improves the energy efficiency of industrial systems [56]. This development led to the application of the SysML modeling language, with the consensus approach contributing to the harmonization of model development [57]. The Munich Agile MBSE Concept (MAGIC) [58] established an agile approach to model-based systems engineering that covers the entire product lifecycle and enables flexible adaptation to changing requirements. In practical applications, this concept has been used, for example, for web-based 3D factory visualizations [59] and has been tested in the development of automated driving functions, such as automated parking. valet [60]. With the further development of the MAGIC concept [61], Salehi positioned an agile MBSE approach as a forward-looking tool for the development of digital products throughout their entire life cycle. This methodology has also been applied to new mobility concepts, such as the use of hydrogen internal combustion engines (HICE) in busses [62] and urban air mobility systems [63]. In addition, Salehi investigated the integration of blockchain technologies into software and system engineering processes [64] and developed security assessments for AI-based systems, particularly in the field of reinforcement learning [65]. The increasing importance of digital twins led to further applications of the MAGIC approach in the development of automated robot-based systems [66] and in the holistic collection of vehicle data through on-board diagnostic systems [67]. In addition, a systematic approach was proposed to generate synthetic data sets from CAD data for CNC production environments to expand the database for AI applications [68]. The use of blockchain technologies in product lifecycle management systems has also been shown to be groundbreaking in ensuring data integrity [69]. In recent work, Salehi focused on the application of system engineering in the construction of three-dimensional environments for autonomous vehicle architectures [70]. In addition, a parametric holistic CAD approach was presented for the development of urban air mobility concepts [71]. The integration of the Munich agile MBSE concept into industrial contexts, particularly with regard to centralized vehicle architectures, marked a further step forward [72]. Finally, the combination of blockchain-based vehicle data architectures with industrial processes led to a new level of data security and traceability in product development [73]. Overall, this work demonstrates a coherent research program that documents the evolution from classic CAD design to an agile, model-based, and system-driven development paradigm. The combination of parametric CAD, MBSE, IoT, and blockchain technologies in integrated lifecycle management illustrates how digital transformation and systems engineering can merge in industry to make future products more efficient, safer, and more sustainable.